\documentclass[conference,usletter]{IEEEtran}
\usepackage{times,amsmath,amssymb}
\usepackage{slashbox}
\usepackage{graphicx}
\usepackage{xcolor}
\usepackage{algorithm2e}
\usepackage{algorithmicx}
\usepackage{setspace}
\usepackage[export]{adjustbox}
\usepackage{multirow}
\usepackage[noend]{algpseudocode}
\usepackage{array,multirow}
\usepackage{makecell}

\let\mc=\multicolumn
\let\mr=\multirow
\let\cl=\cline

\newtheorem{defn}{Definition}

\DeclareMathOperator*{\argminA}{arg\,min}

\begin{document}

\title{Multilinear Compressive Learning}
\author{\IEEEauthorblockN{Dat Thanh Tran\IEEEauthorrefmark{1}, Mehmet Yama\c{c}\IEEEauthorrefmark{1}, Aysen Degerli\IEEEauthorrefmark{1}, Moncef Gabbouj\IEEEauthorrefmark{1}, Alexandros Iosifidis\IEEEauthorrefmark{2}}
\IEEEauthorblockA{\IEEEauthorrefmark{1}Department of Computing Sciences, Tampere University, Tampere, Finland\\
\IEEEauthorrefmark{2}Department of Engineering, Aarhus University, Aarhus, Denmark\\
Email:\{thanh.tran,mehmet.yamac, aysen.degerli, moncef.gabbouj\}@tuni.fi, alexandros.iosifidis@eng.au.dk}\\

}

\maketitle

\begin{abstract}
Compressive Learning is an emerging topic that combines signal acquisition via compressive sensing and machine learning to perform inference tasks directly on a small number of measurements. Many data modalities naturally have a multi-dimensional or tensorial format, with each dimension or tensor mode representing different features such as the spatial and temporal information in video sequences or the spatial and spectral information in hyperspectral images. However, in existing compressive learning frameworks, the compressive sensing component utilizes either random or learned linear projection on the vectorized signal to perform signal acquisition, thus discarding the multi-dimensional structure of the signals. In this paper, we propose Multilinear Compressive Learning, a framework that takes into account the tensorial nature of multi-dimensional signals in the acquisition step and builds the subsequent inference model on the structurally sensed measurements. Our theoretical complexity analysis shows that the proposed framework is more efficient compared to its vector-based counterpart in both memory and computation requirement. With extensive experiments, we also empirically show that our Multilinear Compressive Learning framework outperforms the vector-based framework in object classification and face recognition tasks, and scales favorably when the dimensionalities of the original signals increase, making it highly efficient for high-dimensional multi-dimensional signals.
\end{abstract}

\section{Introduction}\label{S:Intro}

The classical sample-based signal acquisition and manipulation approach usually involve separate steps of signal sensing, compression, storing or transmitting, then the reconstruction. This approach requires the signal to be sampled above the Nyquist rate in order to ensure high-fidelity reconstruction. Since the existence of spatial-multiplexing cameras, over the past decade, Compressive Sensing (CS) \cite{candes2008introduction} has become an efficient and a prominent approach for signal acquisition at sub-Nyquist rates, combining the sensing and compression step at the hardware level. This is due to the assumption that the signal often possesses specific structures that exhibit sparse or compressible representation in some basis, thus, can be sensed at a lower rate than the Nyquist rate but still allows almost perfect reconstruction \cite{candes2006stable, donoho2006compressed}. In fact, many data modalities that we operate on are often sparse or compressible. For example, smooth signals are compressible in the Fourier domain or subsequent frames in a video are piecewise smooth, thus compressible in a wavelet domain. With the efficient realization at the hardware level such as the popular Single Pixel Camera, CS becomes an efficient signal acquisition framework, however, making the signal manipulation an intimidating task. Indeed, over the past decade, since reversing the signal to its original domain is often considered the necessary step for signal manipulation, a significant amount of works have been dedicated to signal reconstruction, giving certain insights and theoretical guarantees for the successful recovery of the signal from compressively sensed measurements \cite{candes2006stable, candes2008introduction, donoho2006compressed}. 

While signal recovery plays a major role in some sensing applications such as image acquisition for visual purposes, there are many scenarios in which the primary objective is the detection of certain patterns or inferring some properties in the acquired signal. For example, in many radar applications, one is often interested in anomaly patterns in the measurements, rather than signal recovery. Moreover, in certain applications \cite{mohassel2017secureml, hesamifard2017cryptodl}, signal reconstruction is undesirable since the step can potentially disclose private information, leading to the infringement of data protection legislation. These scenarios naturally led to the emergence of Compressive Learning (CL) concept \cite{calderbank2012finding, davenport2007smashed, davenport2010signal, reboredo2013compressive} in which the inference system is built on top of the compressively sensed measurements without the explicit reconstruction step. While the amount of literature in CL is rather insignificant compared to signal reconstruction in CS, different attempts have been made to modify the sensing component in accordance with the learning task \cite{baheti2008adaptive,reboredo2013projections}, to extract discriminative features \cite{davenport2007smashed, lohit2015reconstruction} from the randomly sensed measurements or to jointly optimize the sensing matrix \cite{adler2016compressed, lohit2016direct} and the subsequent inference system. Improvements to different components of CL pipeline have been proposed, however, existing frameworks utilize the same compressive acquisition step that performs a linear projection of the vectorized data, thereby operating on the vector-based measurements and thus losing the tensorial structure in the measurements of multi-dimensional data. 

In fact, many data modalities naturally possess the tensorial format such as color images, videos or multivariate time-series. The multi-dimensional representation naturally reflects the semantic differences inherent in different dimensions or tensor modes. For example, the spatial and temporal dimensions in a video or the spatial and the spectral dimensions in hyperspectral images represent two different concepts, having different properties. Thus by exploiting this natural form of the signals and considering the semantic differences between different dimensions, many tensor-based signal processing, and learning algorithms have shown its superiority over the vector-based approach, which simply operates on the vectorized data \cite{nion2010tensor, miwakeichi2004decomposing, dunlavy2011multilinear, tran2017tensor, tran2018improving, cichocki2015tensor, malgouyres2018multilinear}. Indeed, tensor representation and its associated mathematical operations and properties have found various applications in the Machine Learning community. For example, in multivariate time-series analysis, the multilinear projection was utilized in \cite{tran2017tensor, tran2018temporal} to model the dependencies between data points along the feature and temporal dimension separately. Several multilinear regression \cite{youd2002revised, zhao2013higher} or discriminant models \cite{li2014multilinear, tran2017multilinear} have been developed to replace their linear counterparts, with improved performance. In neural network literature, multilinear techniques have been employed to compress pre-trained networks \cite{denton2014exploiting, jaderberg2014speeding, lebedev2014speeding}, or to construct novel neural network architectures \cite{tran2018improving, yang2017tensor, tran2018temporal}.

It is worth noting that CS plays an important role in many applications that involves high-dimensional tensor signals because the standard point-based signal acquisition is both memory and computationally intensive. Representative examples include Hyperspectral Compressive Imaging (HCI), Synthetic Aperture Radar (SAR) imaging, Magnetic Resonance Imaging (MRI) or Computer Tomography (CT). Therefore, the tensor-based approach has also found its place in CS, also known as Multi-dimensional Compressive Sensing (MCS) \cite{caiafa2013multidimensional}, which replaces the linear sensing and reconstruction model with multilinear one. Similar to vector-based CS, thereupon simply referred to as CS, the majority of efforts in MCS are dedicated to constructing multilinear models that induce sparse representation along each tensor mode with respect to a set of bases. For example, the adoption of sparse Tucker representation and the Kronecker sensing scheme in MRI allows computationally efficient signal recovery with very low Peak Signal to Noise Ratio (PSNR) \cite{caiafa2013multidimensional, yu2014multidimensional}. In addition, the availability of optical implementations of separable sensing operators such as \cite{robucci2008compressive} naturally enables MCS, significantly reducing the amount of data collection and reconstruction cost. 

While multilinear models have been successfully applied in Compressive Sensing and Machine Learning, to the best of our knowledge, we have not seen their utilization in Compressive Learning, which is the joint framework combining CS and ML. In this paper, in order to leverage the multi-dimensional structure in many data modalities, we propose Multilinear Compressive Learning framework, which adopts a multilinear sensing operator and a neural network classifier that is designed to utilize the multi-dimensional structure-preserving compressed measurements. The contribution of this paper is as follows:

\begin{itemize}
\item We propose Multilinear Compressive Learning (MCL), a novel CL framework that consists of a multilinear sensing module, a multilinear feature synthesis component, both taking into account the multi-dimensional property of the signals, and a task-specific neural network. The multilinear sensing module compressively senses along each separate mode of the original tensor signal, producing structurally encoded measurements. Similarly, the feature synthesis component performs the feature learning steps separately along each mode of the compressed measurements, producing inputs to the subsequent task-specific neural network which has the structure depending on the inference problem. 

\item We show both theoretically and empirically that the proposed MCL framework is highly cost-effective in terms of memory and computational complexity. In addition, theoretical analysis and experimental results also indicate that our framework scales well when the dimensionalities of the original signal increases, making it highly efficient for high-dimensional tensor signals.  

\item We conduct extensive experiments in object classification and face recognition tasks to validate the performance of our framework in comparison with its vector-based counterpart. Besides, the effect of different components and hyperparameters in the proposed framework were also empirically analyzed. 

\item We publicly provide our implementation of the experiments reported in this paper to facilitate future research. By following our detailed instructions on how to set up the software environment, all experiment results can be reproduced in just one line of code. \footnote{https://github.com/viebboy/MultilinearCompressiveLearningFramework}

\end{itemize}

The remainder of the paper is organized as follows: in Section 2, we review the background information in Compressive Sensing, Multi-dimensional Compressive Sensing and Compressive Learning. In Section 3, the detailed description of the proposed Multilinear Compressive Learning framework is given. Complexity analysis and comparison with the vector-based framework are also given in Section 3. In Section 4, we provide details of our experiment protocols and quantitative analysis of different experiment configurations. Section 5 concludes our work with possible future research directions. 

\section{Related Work}

\subsection{Notation}
In this paper, we denote scalar values by either lower-case or upper-case characters $(x, y, X, Y \dots)$, vectors by lower-case bold-face characters $(\mathbf{x}, \mathbf{y}, \dots)$, matrices by upper-case or Greek bold-face characters $(\mathbf{A}, \mathbf{B}, \mathbf{\Phi}, \dots)$ and tensor as calligraphic capitals $(\mathcal{X}, \mathcal{Y}, \dots)$. A tensor with $K$ modes and dimension $I_{k}$ in the mode-$k$ is represented as $\mathcal{X} \in \mathbb{R}^{I_1 \times I_2 \times \dots \times I_K}$. The entry in the $i_k$th index in mode-$k$ for $k=1,\dots, K$ is denoted as $\mathcal{X}_{i_1,i_2,\dots,i_K}$. In addition, $vec(\mathcal{X})$ denotes the vectorization operation that rearranges elements in $\mathcal{X}$ to the vector representation. 

\begin{defn}[The Kronecker Product]\label{def1}
The Kronecker product between two matrices $\mathbf{A} \in \mathbb{R}^{M\times N}$ and $\mathbf{B} \in \mathbb{R}^{P\times Q}$ is denoted as $\mathbf{A} \otimes \mathbf{B}$ having dimension $MP \times NQ$, is defined by:
\begin{equation}\label{kronecker}
\mathbf{A} \otimes \mathbf{B} = \begin{bmatrix} \mathbf{A}_{11} \mathbf{B} &  \dots & \mathbf{A}_{1N} \mathbf{B} \\ \vdots & \ddots & \vdots \\ \mathbf{A}_{M1} \mathbf{B} & \hdots & \mathbf{A}_{MN} \mathbf{B} \end{bmatrix}
\end{equation}
\end{defn}

\begin{defn}[Mode-$n$ Product]\label{def2}
	The mode-$k$ product between a tensor $\mathcal{X}=[x_{i_1},\dots , x_{i_N}] \in  \mathbb{R}^{I_1 \times \dots I_N}$ and a matrix $\mathbf{W}\in \mathbb{R}^{J_{n}\times I_n}$ is another tensor of size $I_1\times \dots \times J_{n}\times \dots \times I_N$ and denoted by $\mathcal{X} \times_{n} \mathbf{W}$. The element of $\mathcal{X} \times_{n} \mathbf{W}$ is defined as $[\mathcal{X}\times_{n}\mathbf{W}]_{i_1, \dots , i_{n-1}, j_n, i_{n+1},\dots, i_N}=\sum_{i_n=1}^{I_N}[\mathcal{X}]_{i_1,\dots,i_{n-1},i_n,\dots, i_N}[\mathbf{W}]_{j_n,i_n}$.
\end{defn}

The following relationship between the Kronecker product and $n$-mode product is the cornerstone in MCS:
\begin{equation}\label{property1}
\mathcal{Y} = \mathcal{X} \times_1 \mathbf{W}_1 \times \dots \times_N \mathbf{W}_N
\end{equation}
can be written as
\begin{equation}\label{property2}
\mathbf{y} = (\mathbf{W}_1 \otimes \dots \otimes \mathbf{W}_N) \mathbf{x}
\end{equation}
where $\mathbf{y} = vec(\mathcal{Y})$ and $\mathbf{x} = vec(\mathcal{X})$

\subsection{Compressive Sensing}

Compressive Sensing (CS) \cite{candes2008introduction} is a signal acquisition and manipulation paradigm that performs simultaneous sensing and compression on the hardware level, leading to large reduction in computation cost and the number of measurements. The signal $\mathbf{y}$ working under CS is assumed to have a sparse or compressible representation $\mathbf{x}$ in some basis or dictionary $\mathbf{\Psi} \in \mathbb{R}^{I\times I}$, that is:

\begin{equation}\label{eq1}
\mathbf{y} = \mathbf{\Psi} \mathbf{x} \quad\mathrm{with}\quad \|\mathbf{x}\|_0 \leq K \quad\mathrm{and}\quad  K \ll I
\end{equation}
where $\|\mathbf{x}\|_0$ denotes the number of non-zero entries in $\mathbf{x}$. While the dictionary presented in Eq. (\ref{eq1}) is complete, i.e., the number of columns in $\mathbf{\Psi}$ is equal to the signal dimension $I$, we should note that signal models with over-complete dictionaries can also work, i.e., $\mathbf{\Psi} \in \mathbb{R}^{J\times I}$ with some modifications \cite{aharon2006k}. 

With the assumption on the sparsity, CS performs the linear sensing step using the sensing operator $\mathbf{\Phi} \in \mathbb{R}^{M \times I}$, acquiring a small number of measurements $\mathbf{z} \in \mathbb{R}^{M}$ with $M<I$, from analog signal $\mathbf{y}$:

\begin{equation}\label{eq3}
\mathbf{z} = \mathbf{\Phi} \mathbf{y}
\end{equation}

Eq. (\ref{eq3}) represents both the sensing and compression step that can be efficiently implemented at the sensor level. Thus, what we obtain from CS sensors is a limited number of measurements $\mathbf{z}$ that is used for other processing steps. By combining Eq. (\ref{eq1}, and \ref{eq3}), the CS model is usually expressed as:

\begin{equation}\label{eq4}
\mathbf{z} = \mathbf{\Phi} \mathbf{\Psi} \mathbf{x} \quad\mathrm{with}\quad \|\mathbf{x}\|_0 \leq K \quad\mathrm{and}\quad  K \ll I
\end{equation}

In some applications, we are interested in recovering the signal $\mathbf{y}$ from $\mathbf{z}$. This involves developing theoretical properties and algorithms to determine the sensing operator $\mathbf{\Phi}$, the dictionary or basis $\mathbf{\Psi}$, and the number of nonzero coefficients $K$ in order to ensure that the reconstruction is unique, and of high-fidelity \cite{candes2006stable, donoho2003optimally, donoho2006compressed}. The reconstruction of $\mathbf{y}$ is often posed as finding the sparsest solution of the under-determined linear system \cite{tropp2010computational}, particularly:

%It has been proven that with the existence of the sparse signal model in Eq. (\ref{eq1}), the number of measurements required for near-perfect signal recovery is $O(K\log(I/K))$. Besides, several techniques have been developed to provide the conditions for unique reconstruction, which are usually expressed via two metrics on $\mathbf{\Phi} \mathbf{\Psi}$: \textit{spark} which measures the minimum number of linearly dependent columns, and \textit{coherence} which is the maximum normalized absolute inner product between any two columns. Generally, $\mathbf{\Phi} \mathbf{\Psi}$ with either large spark ($\mathrm{spark}(\mathbf{\Phi} \mathbf{\Psi} ) > 2K^{35}$) or low coherence ($\mu (\mathbf{\Phi} \mathbf{\Psi}) < 1/(2K-1)^{35}$) can ensure uniqueness. 

\begin{equation}\label{eq5}
\argminA_{\mathbf{x}} \|\mathbf{x}\|_0 \quad\mathrm{s.t}\quad \|\mathbf{z} - \mathbf{\Phi} \mathbf{\Psi} \mathbf{x}\|_2 \leq \epsilon
\end{equation}
where $\epsilon$ is a small constant specifying the amount of residual error allowed in the approximation. A large body of research has been dedicated to solve the problem in Eq. (\ref{eq5}) and its variants with two main approaches: \textit{basis pursuit} (BP) which transforms Eq. (\ref{eq5}) to a convex one to be solved by linear programming \cite{chen2001atomic} or second-order cones programs \cite{candes2006stable}, and \textit{matching pursuit} (MP), a class of greedy algorithms, which iteratively refines the solution to the sparsest \cite{tropp2004greed, tropp2007signal}. Both BP and MP algorithms are computationally intensive when the number of elements in $\mathbf{y}$ is big, especially in the case of multi-dimensional signals.  

\subsection{Multi-dimensional Compressive Sensing}

Given a multi-dimensional signal $\mathcal{Y} \in \mathbb{R}^{I_1 \times \dots \times I_N}$, a direct application of the sparse representation in Eq. (\ref{eq1}) requires vectorizing $\mathbf{y} = vec(\mathcal{Y})$ and the calculations on $\mathbf{\Phi} \mathbf{\Psi} \in \mathbb{R}^{M \times (I_1 \dots I_N)}$, which is a very big matrix with the number of elements scales exponentially with $N$. Instead of assuming $vec(\mathcal{Y})$ is sparse in some basis or dictionary, MCS adopts a sparse Tucker model \cite{de2000multilinear} as follows:

\begin{equation}\label{eq6}
\mathcal{Y} = \mathcal{X} \times_{1} \mathbf{\Psi}_1 \times \dots \times_N \mathbf{\Psi}_N
\end{equation}
which assumes that the signal $\mathcal{Y}$ is sparse with respect to a set of bases or dictionaries $\mathbf{\Psi}_n, n=1, \dots, N$. Since in some cases, the sensing step can be taken in a multilinear way, i.e., by using a set of linear operators along each mode separately, also known as separable sensing operators:

\begin{equation}\label{eq7}
\mathcal{Z} = \mathcal{Y} \times_1 \mathbf{\Phi}_1 \times \dots \times_N \mathbf{\Phi}_N
\end{equation}
that allows us to obtain the measurements $\mathcal{Z}$ with retained multi-dimensional structure. From Eq. (\ref{property1}, \ref{property2}, \ref{eq6} and \ref{eq7}), the MCS model is often expressed as:

\begin{equation}\label{eq8}
\mathbf{z} = (\mathbf{B}_1 \otimes \dots \otimes \mathbf{B}_N) \mathbf{x} \quad\mathrm{with}\quad \|\mathbf{x}\|_0 \leq K
\end{equation}
where $\mathbf{z} = vec(\mathcal{Z})$, and $\mathbf{B}_n = \mathbf{\Phi}_n \mathbf{\Psi}_n$ ($n=1, \dots, N$). The formulation in Eq. (\ref{eq8}) is also known as Kronecker CS \cite{duarte2012kronecker}.

Since MCS can be expressed in the vector form, the existing algorithms and theoretical bounds for vector-based CS have also been extended for MCS. Representative examples include Kronecker OMP and its tensor block-sparsity extension \cite{caiafa2013computing} that improves the computation significantly. It is worth noting that by adopting a multilinear structure, MCS operates with a set of smaller sensing and dictionaries, requiring much lower memory and computation compared to the vectorization approach \cite{caiafa2013multidimensional}. 

\subsection{Compressive Learning}
\begin{figure*}[t]
	\centering
	\includegraphics[width=0.95\textwidth]{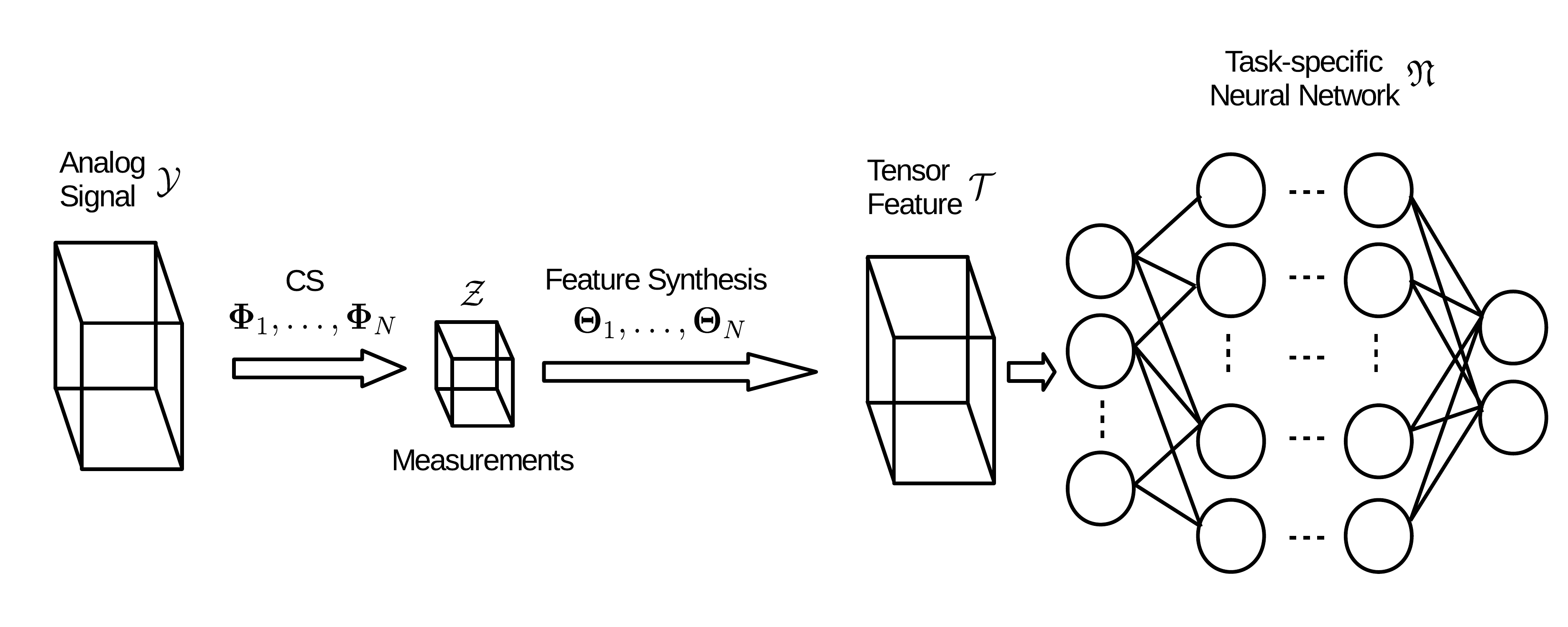}
	\caption{Illustration of the proposed Multilinear Compressive Learning framework}\label{figure1}
\end{figure*}

The idea of learning directly from the compressed measurements dates back to the early work of \cite{davenport2007smashed} in which the authors proposed a framework termed \textit{compressive classification} which introduces the concept of \textit{smashed filters} and operates directly on the compressive measurements without reconstruction as the first proxy step. The result in \cite{davenport2007smashed} was subsequently strengthened in \cite{baraniuk2009random} showing that when sufficiently large random sensing matrix is used, it can capture the structure of the data manifold. Later, further extensions that extract discriminative features from compressive measurements for activity recognition \cite{kulkarni2012recurrence, kulkarni2016reconstruction} or face recognition \cite{lohit2015reconstruction} have also been proposed.  

The concept of CL was introduced in \cite{calderbank2012finding}, which provides theoretical analysis illustrating that learning machines can be built directly in the compressed domain. Particularly, given certain conditions of the sensing matrix $\mathbf{\Phi}$, the performance of a linear Support Vector Machine (SVM) trained on compressed measurements is as good as the best linear threshold classifier trained on the original signal $\mathbf{y}$. Later, for compressive learning of signals described by a Gaussian Mixture Model, asymptotic behavior of the upper-bound \cite{reboredo2013compressive} and its extension \cite{reboredo2013projections} to learn the sensing matrix were also derived. 

The idea of jointly optimizing the sensing matrix with the classifier was also adopted in \cite{baheti2008adaptive} in which the authors proposed an adaptive version of \textit{feature-specific imaging} system to learn an optimal sensing matrix based on past measurements. With the advances in computing hardware and stochastic optimization techniques, end-to-end CL system was proposed in \cite{adler2016compressed}, and several follow-up extensions and applications \cite{hollis2018compressed, deugerli2018compressively, xu2019compressed}, indicating the superior performance when simultaneously optimizing the sensing component and the classifier via task-specific data. Our work is closely related to the end-to-end CL system in \cite{adler2016compressed} in that we also optimize the CL system via stochastic optimization in an end-to-end manner. Different from \cite{adler2016compressed}, our proposed framework efficiently utilizes the tensor structure inherent in many types of signals, thus outperforming the approach in \cite{adler2016compressed} in both inference performance and computational efficiency.

\section{Multilinear Compressive Learning Framework}

In this Section, we first give our description of the proposed Multilinear Compressive Learning (MCL) framework that operates directly on the tensor representation of the signals. Then, the initialization scheme and optimization procedures of the proposed framework is discussed. Lastly, theoretical analysis of the framework's complexity in comparison with its vector-based counterpart is provided.

\subsection{Motivation}

In order to model the multi-dimensional structure in the signal of interest, we assume that the discriminative structure in $\mathcal{Y} \in \mathbb{R}^{I_1 \times \dots \times I_N}$ can be captured in a lower-dimensional multilinear subspace $\mathfrak{F} \subset \mathbb{R}^{J_1 \times \dots \times J_N}$ of $\mathbb{R}^{I_1 \times \dots \times I_N}$ with ($J_n < I_n, \forall n=1, \dots, N$):

\begin{equation}\label{eq9}
\mathcal{Y} = \bar{\mathcal{X}} \times_1 \bar{\mathbf{\Psi}}_1 \times \dots \times_N \bar{\mathbf{\Psi}}_N 
\end{equation}
where $\bar{\mathbf{\Psi}}_n \in \mathbb{R}^{I_n \times J_n}, \forall n=1, \dots, N$ denotes the factor matrices and $\bar{\mathcal{X}} \in \mathbb{R}^{J_1 \times \dots \times J_N}$ is the signal representation in this multilinear subspace. 

Here we should note that although Eq. (\ref{eq9}) in our framework and Eq. (\ref{eq6}) in MCS look similar in its mathematical form, the assumption and motivation are different. The objective in MCS is to reconstruct the signal $\mathcal{Y}$ by assuming the existence of the set of sparsifying dictionaries or bases $\mathbf{\Psi}_n$ and optimizing $\mathbf{\Psi}_n$ to induce the sparsest $\mathcal{X}$. Since our objective is to learn a classification or regression model, we make no assumption or constraint on the sparsity of $\bar{\mathcal{X}}$ but assume that the factorization in Eq. (\ref{eq9}) can lead to a tensor subspace $\mathfrak{F}$ in which the representation $\bar{\mathcal{X}}$ is discriminative or meaningful for the learning problem. 

As mentioned in the previous Section, in some applications, the measurements can be taken in a multilinear fashion, with different linear sensing operators operating along different tensor modes, i.e., separable sensing operators, we obtain the measurements $\mathcal{Z}$ from the following sensing equation:

\begin{equation}\label{eq10}
\mathcal{Z} = \mathcal{Y} \times_ 1 \mathbf{\Phi}_1 \times \dots \times_N \mathbf{\Phi}_N
\end{equation} 
where $\mathbf{\Phi}_n \in \mathbb{R}^{M_n \times I_n}$ ($n=1, \dots, N$) represent the sensing matrices of those linear operators. 

In cases where the measurements of the multi-dimensional signals are taken in a vector-based fashion, i.e., the following sensing model:

\begin{equation}\label{eq11}
\mathbf{z} = \mathbf{\Phi} vec(\mathcal{Y})
\end{equation}
with a single sensing operator $\mathbf{\Phi} \in \mathbb{R}^{M \times I_1 \ast \dots \ast I_N}$, we can still enforce a structure-preserving sensing operation similar to the multilinear sensing scheme in Eq. (\ref{eq10}) by setting:

\begin{equation}\label{eq12}
\mathbf{\Phi} = \mathbf{\Phi}_1 \otimes \dots \otimes \mathbf{\Phi}_N 
\end{equation}
to obtain $\mathcal{Z}$ in Eq. (\ref{eq10}) from $\mathbf{z}$ in Eq. (\ref{eq11}).

Combining Eq. (\ref{eq9} and \ref{eq10}), we can express our measurements $\mathcal{Z}$ as:

\begin{equation}\label{eq13}
\mathcal{Z} = \bar{\mathcal{X}} \times_1 (\mathbf{\Phi}_1 \bar{\mathbf{\Psi}}_1) \times \dots \times_N (\mathbf{\Phi}_N \bar{\mathbf{\Psi}}_N)
\end{equation}

By setting the sensing matrices $\mathbf{\Phi}_n$ to be pseudo-inverse of $\bar{\mathbf{\Psi}}_n$ for all $n=1, \dots, N$, we obtain the measurements $\mathcal{Z}$ that lie in the discriminative-induced tensor subspace $\mathfrak{F}$ mentioned previously.

\subsection{Design}

Figure \ref{figure1} illustrates our proposed MCL framework which consists of the following components:

\begin{itemize}
\item CS component: the data acquisition step of the multi-dimensional signals is done via separable linear sensing operators $\mathbf{\Phi}_n, n=1, \dots, N$. As mentioned previously, in cases where the actual hardware implementation only allows vector-based sensing scheme, Eq. (\ref{eq12}) allows the simulation of this multilinear sensing step. This component produces measurements $\mathcal{Z}$ with encoded tensor structure, having the same number of tensor modes ($N$) as the original signal.  

\item Feature Synthesis (FS) component: from $\mathcal{Z}$, this step performs feature extraction along $N$ modes of the measurements $\mathcal{Z}$ with the set of learnable matrices $\mathbf{\Theta}_n$. Since the measurements typically have many fewer elements compared to the original signal $\mathcal{Y}$, the FS component expands the dimensions of $\mathcal{Z}$, allowing better separability between the sensed signals from different classes in a higher multi-dimensional space that is found through optimization. While the sensing step performs linear interpolations for computational efficiency, the FS component can be either multilinear or nonlinear transformations. A typical nonlinear transformation step is to perform zero-thresholding, i.e., ReLU, on $\mathcal{Z}$ before multiplying with $\mathbf{\Theta}_n, n=1\dots, N$, i.e., $\mathrm{ReLU}(\mathcal{Z}) \times_1 \mathbf{\Theta}_1 \times \dots \times_N \mathbf{\Theta}_N$. In applications which require the transmission of $\mathcal{Z}$ to be analyzed, this simple thresholding step can, before transmission, increase the compression rate by sparsifying the encoded signal and discarding the sign bits. While nonlinearity is often considered beneficial for neural networks, adding the thresholding step as described above further restricts the information retained in a limited number of measurements $\mathcal{Z}$, thus, adversely affects the inference system. In the Experiments Section, we provide empirical analysis on the effect of nonlinearity towards the inference tasks at different measurement rates. Here we should note that while our FS component resembles the reprojection step in the vector-based framework \cite{adler2016compressed}, our FS and CS components have different weights ($\mathbf{\Theta}_n$ and $\mathbf{\Phi}_n$, $n=1, \dots, N$) and the dimensionality of the tensor feature $\mathcal{T}$ produced by FS component is task-dependent, and is not constrained to that of the original signal. 

\item Task-specific Neural Network $\mathfrak{N}$: from the tensor representation $\mathcal{T}$ produced by FS step, a neural network with task-dependent architecture is built on top to generate the regression or classification outputs. For example, when analyzing visual data, the $\mathfrak{N}$ can be a Convolutional Neural Network (CNN) in case of static images or a Convolutional Recurrent Neural Network in case of videos. In CS applications that involve distributed arrays of sensors that continuously collect data, specific architectures for time-series analysis such as Long-Short Term Memory Network should be considered for $\mathfrak{N}$. Here we should note that the size of $\mathcal{T}$ is also task-dependent and should match with the neural network component. For example, in object detection and localization task, it is desirable to keep the spatial aspect ratio of $\mathcal{T}$ similar to $\mathcal{Y}$ to allow precise localization. 
\end{itemize}

\subsection{Optimization}

\begin{table*}[t!]
	\begin{center}
		\caption{Complexity of the proposed MCL framework and vector-based framework \cite{adler2016compressed}}\label{t1}
		\resizebox{0.95\textwidth}{!}{
			\begin{tabular}{|c|c|c|}\hline
				& Our 								& Vector \cite{adler2016compressed}			\\ \hline
				Memory		& $O(2\sum_{n=1}^{N}I_n \ast M_n)$	& $O(\prod_{n=1}^{N} I_n \ast M_n)$	\\ \hline		
				Computation	&  $O\big(\sum_{n=1}^{N} \big(\prod_{p=1}^{n}M_p \ast \prod_{k=n}^{N}I_k\big) + \sum_{n=1}^{N} \big(\prod_{p=1}^{n}I_p \ast  \prod_{k=n}^{N} M_k\big))$ & $O(2\prod_{n=1}^{N} I_n \ast M_n)$ \\ \hline
			\end{tabular}
		}
	\end{center}
\end{table*}

In our proposed MCL framework, we aim to optimize all three components, i.e., $\mathbf{\Phi}_n$, $\mathbf{\Theta}_n$ and $\mathfrak{N}$, with respect to the inference task. A simple and straightforward approach is to consider all components in this framework as a single computation graph, then randomly initialize the weights according to some popular initialization scheme \cite{glorot2010understanding, he2016identity} and perform stochastic gradient descend on this graph with respect to the loss function defined by the learning task. However, this approach does not take into account any existing domain knowledge of each component that we have. 

As mentioned in Section III.A, with the assumption of the existence of a tensor subspace $\mathfrak{F}$ and the factorization in Eq. (\ref{eq9}), the sensing matrix $\mathbf{\Phi}_n$ in the CS component can be initialized equal to the pseudo-inverse of $\bar{\mathbf{\Psi}}_n$ for all $n=1,\dots, N$ to obtain initial $\mathcal{Z}$ that are discriminative or meaningful. There have been several algorithms proposed to learn the factorization in Eq. (\ref{eq9}) with respect to different criteria such as the multi-class discriminant \cite{li2014multilinear}, class-specific discriminant \cite{tran2017multilinear}, max-margin \cite{wu2013supervised} or Tucker Decomposition with non-negative constraint \cite{kim2007nonnegative}. 

In a general setting, we propose to apply Higher Order Singular Value Decomposition (HOSVD) \cite{de2000multilinear} and initialize $\mathbf{\Phi}_n$ with the left singular vectors that correspond to the largest singular values in mode $n$. The sensing matrices are then adjusted together with other components during the stochastic optimization process. This initialization scheme resembles the one proposed for vector-based CL framework which utilizes Principal Component Analysis (PCA). In a general case where one has no prior knowledge on the structure of $\mathfrak{F}$, a transformation that retains the most energy in the signal such as PCA or HOSVD is a popular choice when reducing dimensionalities of the signal. While for higher-order data, HOSVD only provides a quasi-optimal condition for data reconstruction in the least-square sense \cite{grasedyck2013literature}, since our objective is to make inferences, this initialization scheme works well as indicated in our Experiments Section.

With the aforementioned initialization scheme of CS component for a general setting, it is natural to also initialize $\mathbf{\Theta}_n$ in FS component with the right singular vectors corresponding to the largest singular values in mode $n$ of the training data. With this initialization of $\mathbf{\Theta}_n$, during the initial forward steps in stochastic gradient descent, the FS component produces an approximate version of $\mathcal{Y}$, and in cases where a classifier $\mathfrak{C}$ pre-trained on $\mathcal{Y}$ or its approximated version $\hat{\mathcal{Y}}$ exists, the weights of neural network $\mathfrak{N}$ can be initialized with that of $\mathfrak{C}$. It is worth noting that the reprojection step in the vector-based framework in \cite{adler2016compressed} shares the weights with the sensing matrices, performing inexplicit signal reconstruction while we have different sensing $\mathbf{\Phi}_n, n=1, \dots, N$ and feature extraction $\mathbf{\Theta}_n, n=1, \dots, N$ weights. Since the vector-based framework involves large sensing and reprojection matrices, from the optimization point of view, enforcing shared weights might be essential in their framework to reduce overfitting as indicated by their empirical results. 

After performing the aforementioned initialization steps, all three components in our MCL framework are optimized using Stochastic Gradient Descent method. It is worth noting that above initialization scheme for CS and FS component is proposed in a generic setting, which can serve as a good starting point. In cases where certain properties of the tensor subspace $\mathfrak{F}$ or the tensor feature $\mathcal{T}$ are known to improve the learning task, one might adopt a different initialization strategy for CS and FS components to induce such properties. 

\subsection{Complexity Analysis}

Since the complexity of the neural network component $\mathfrak{N}$ varies with the choice of the architecture, we will estimate the theoretical complexity for the CS and FS component and make comparison with the vector-based framework \cite{adler2016compressed}. Let $\mathbb{R}^{I_1 \times \dots \times I_N}$ and $\mathbb{R}^{M_1 \times \dots \times M_N}$ denote the dimensionality of the original signal $\mathcal{Y}$ and its measurements $\mathcal{Z}$, respectively. In addition, to compare with the vector-based framework, we also assume that the dimensionality of the feature $\mathcal{T}$ is also $\mathbb{R}^{I_1 \times \dots \times I_N}$. Thus, $\mathbf{\Phi}_n$ belongs to $\mathbb{R}^{M_n \times I_n}$ and $\mathbf{\Theta}_n$ belongs to $\mathbb{R}^{I_n \times M_n}$ for $n=1, \dots, N$ in our CS and FS component, while in \cite{adler2016compressed}, the sensing matrix $\mathbf{\Phi}$ and the reconstruction matrix $\mathbf{\Phi}^T$ belong to $\mathbb{R}^{(I_1 \ast \dots \ast I_N) \times (M_1 \ast \dots \ast M_N)}$ and $\mathbb{R}^{(M_1 \ast \dots \ast M_N) \times (I_1 \ast \dots \ast I_N)}$, respectively. 

It is clear that the memory complexity of CS and FS component in our MCL framework is $O(2\sum_{n=1}^{N}I_n \ast M_n)$, and that of the vector-based framework is $O(\prod_{n=1}^{N} I_n \ast M_n)$. To see the huge difference between the two frameworks, let us consider 3D MRI image of size $I_1 \times I_2 \times I_3 = 256 \times 256 \times 64$ with the sampling ratio $70\%$, i.e., $M_1 \times M_2 \times M3 = 214 \times 214 \times 64$, the memory complexity in our framework is $O(256\ast214 + 256\ast 214 + 64\ast 64) \approx O(10^5)$ while that of the vector-based framework is $O(256 \ast 214 \ast 256 \ast 214 \ast 64 \ast 64) = O(10^{13})$

Regarding computational complexity of our framework, the CS component performs $\mathcal{Z} = \mathcal{Y} \times_1 \mathbf{\Phi}_1 \times \dots \times_N \mathbf{\Phi}_N$ having complexity of $O\big(\sum_{n=1}^{N} \big(\prod_{p=1}^{n}M_p \ast \prod_{k=n}^{N}I_k\big)\big)$, and the FS component performs $\mathcal{T} = \mathcal{Z} \times_ 1 \mathbf{\Theta}_1 \times \dots \times_N \mathbf{\Theta}_N$ having complexity of $O\big(\sum_{n=1}^{N} \big(\prod_{p=1}^{n}I_p \ast  \prod_{k=n}^{N}M_k\big)\big)$. For the vector-based framework, the sensing step computes $\mathbf{z} = \mathbf{\Phi} vec(\mathcal{Y})$ and reprojection step computes $\mathbf{\Phi}^T \mathbf{z}$, resulting in total complexity of $O(2\prod_{n=1}^{N}I_n \ast M_n)$. With the same 3D MRI example as in the previous paragraph, the total computational complexity of our framework is $O(10^9)$ while that of the vector-based framework is $O(10^{13})$. 

Table \ref{t1} summarizes the complexity of the two frameworks. It is worth noting that by taking into account the multi-dimensional structure of the signal, the proposed framework has both memory and computational complexity several orders of magnitudes lower than its vector-based counterpart.

\section{Experiments}

In this section, we provide a detailed description of our empirical analysis of the proposed MCL framework. We start by describing the datasets and the experiments' protocols that have been used. In the standard set of experiments, we analyze the performance of MCL in comparison with the vector-based framework proposed in \cite{adler2016compressed, zisselman2018compressed}. We further investigate the effect of different components in our framework in the Ablation Study Subsection. 

\subsection{Datasets and Experiment Protocol}

We have conducted experiments on the object classification and face recognition tasks on the following datasets:

\begin{itemize}
\item CIFAR-10 and CIFAR-100: CIFAR dataset \cite{krizhevsky2009learning} is a color (RGB) image dataset for evaluating object recognition task. The dataset consists of $50\mathrm{K}$ images for training and $10\mathrm{K}$ images for testing with resolution $32\times 32$ pixels. CIFAR-10 refers to the $10$-class objection recognition task in which each individual image has a single class label coming from $10$ different categories. Likewise, CIFAR-100 refers to a more fine-grained classification task with each image having a label coming from $100$ different categories. In our experiment, from the training set of CIFAR-10 and CIFAR-100, we randomly selected $5\mathrm{K}$ images for validation purpose and only trained the algorithms on $45\mathrm{K}$ images. 

\item CelebA: CelebA \cite{liu2015faceattributes} is a large-scale face attributes dataset with more than $200\mathrm{K}$ images at different resolutions from more than $10\mathrm{K}$ identities. In our experiment, we used a subset of $100$ identities in this dataset which corresponds to $7063$, $2373$, and $2400$ samples for training, validation, and testing, respectively. In order to evaluate the scalability of our proposed framework, we resized the original images to different set of resolutions, including: $32 \times 32$, $48\times 48$, $64\times 64$, and $80\times 80$ pixels, which are subsequently denoted as CelebA-32, CelebA-48, CelebA-64, and CelebA-80, respectively.

\end{itemize}

In our experiments, two types of network architecture have been employed for the neural network component $\mathfrak{N}$: the AllCNN architecture \cite{springenberg2014striving} and the ResNet architecture \cite{he2015delving}. AllCNN is a simple 9-layer feed-forward architecture which has no max-pooling (pooling is done via convolution with stride more than 1) and no fully-connected layer. ResNet is a 110-layer CNN with residual connections. The exact topologies of AllCNN and ResNet in our experiment can be found in our publicly available implementation\footnote{https://github.com/viebboy/MultilinearCompressiveLearningFramework}.

Since all of the datasets contain RGB images, we followed the implementation proposed in \cite{zisselman2018compressed} for the vector-based framework, which is an extension of \cite{adler2016compressed}, which has 3 different sensing matrices for each of the color channel, and the corresponding reprojection matrices are enforced to share weights with the sensing matrices. The sensing matrices in MCL were initialized with the HOSVD decomposition on the training sets while the sensing matrices in the vector-based framework were initialized with PCA decomposition on the training set. Likewise, the bases obtained from HOSVD and PCA were also used to initialize the FS component in our framework and the reprojection matrices in the vector-based framework. In addition, we also trained the neural network component $\mathfrak{N}$ on uncompressed data with respect to the learning tasks and initialized the classifier in each framework with these pre-trained networks' weights. After the initialization step, both frameworks were trained in an end-to-end manner. 

All algorithms were trained with ADAM optimizer \cite{kingma2014adam} with the following learning rate the schedule $\{10^{-3}, 10^{-4}, 10^{-5}\}$, changing at epoch $80$ and $120$. Each algorithm was trained for $160$ epochs in total. Weight decay coefficient was set to $0.0001$ to regularize all the trainable weights in all experiments. We performed no data preprocessing step, except scaling all the pixel values to $[0, 1]$. In addition, data augmentation was employed by random flipping on the horizontal axis and image shifting within $10\%$ of the spatial dimensions. In all experiments, the final model weights which are used to measure the performance on the test sets, are obtained from the epoch which has the highest validation accuracy.
 
For each experiment configuration, we performed $3$ runs and the mean and standard deviation of test accuracy are reported. 

\subsection{Comparison with the vector-based framework}

\begin{table}[]
	\begin{center}
		\caption{Different configurations of measurements between vector-based framework and our framework. * \textit{Measurement Rate} is calculated with respect to the original signal of size $32\times 32\times 3$}\label{t2}
		\resizebox{\linewidth}{!}{
			\begin{tabular}{|c|c|c|c|}\hline
				
				Type 								& Configuration 	& \#measurements 	& Measurement Rate \\ \hline \hline
				vector \cite{zisselman2018compressed} 	& $256 \times 3$		& $768$				& $0.250$ \\ \hline 
				MCL (our)						& $20\times 19\times 2$	& $760$				& $0.247$ \\ \hline
				MCL (our)						& $28\times 27\times 1$	& $756$				& $0.246$ \\ \hline \hline
				
				vector \cite{zisselman2018compressed} 	& $102 \times 3$		& $306$				& $0.100$ \\ \hline 
				MCL (our)						& $14\times 11\times 2$	& $308$				& $0.100$ \\ \hline
				MCL (our)						& $18\times 17\times 1$	& $306$				& $0.100$ \\ \hline \hline
				
				vector \cite{zisselman2018compressed} 	& $18 \times 3$			& $54$				& $0.018$ \\ \hline 
				MCL (our)						& $9\times 6\times 1$	& $54$				& $0.018$ \\ \hline
				MCL (our)						& $6\times 9\times 1$	& $54$				& $0.018$ \\ \hline

			\end{tabular}
		}
	\end{center}
\end{table}

In order to compare with the vector-based framework in \cite{adler2016compressed, zisselman2018compressed}, we performed experiments on 3 datasets: CIFAR-10, CIFAR-100, and CelebA-32. To compare the performances at different measurement rates, we employed three different measurement values $Z$ for the vector-based framework: $256\times 3=768$, $102\times 3= 306$, and $18\times 3=54$. Here $\times 3$ indicates that the vector-based framework has $3$ different sensing matrices for each color channel. Since we cannot always select the size of the measurements $\mathcal{Z}$ in MCL to match the number of measurements in the vector-based framework, we try to find the configurations of $\mathcal{Z}$ that closely match with the vector-based ones. In addition, with a target number of measurements, there can be more than one configuration of $\mathcal{Z}$ that yields a similar number of measurements. For each measurement value ($768, 102, 54$) in the vector-based framework, we evaluated two different values of $\mathcal{Z}$, particularly, the following sizes of $\mathcal{Z}$ were used: $20\times 19 \times 2 = 760$, $28 \times 27 \times 1= 756$, $14 \times 11 \times 2= 308$, $18 \times 17 \times 1= 306$, $6 \times 9 \times 1 = 54$ and $9 \times 6 \times 1=54$. The measurement configurations are summarized in Table \ref{t2}. 

In order to effectively compare the CS and FS component in MCL with those in \cite{zisselman2018compressed}, two different neural network architectures with different capacities have been used. Table \ref{t3} and \ref{t4} show the accuracy on the test set with AllCNN and ResNet architecture, respectively. The second row of each table shows the performance of the base classifier on the uncompressed data, which we term as \textit{Oracle}. 

It is clear that our proposed framework outperforms the vector-based framework in all compression rates and datasets with both AllCNN and ResNet architecture, except for CIFAR-100 dataset at the lowest measurement rate ($0.018$). The performance gaps between the proposed MCL framework and the vector-based one are huge, with more than $10\%$ differences for the CIFAR datasets at measurement rates $0.25$ and $0.10$. In case of CelebA-32 dataset and at measurement rate $0.246$ (configuration $28\times 27\times 1$), the inference systems learned by our proposed framework even slightly outperform the Oracle setting for both AllCNN and ResNet architecture. 

Although the capacities of AllCNN and ResNet architecture are different, their performances on the uncompressed data are roughly similar. Regarding the effect of two different base classifiers in the two Compressive Learning pipelines, it is clear that the optimal configurations of our framework at each measurement rate are consistent between the two classifiers, i.e., the bold patterns from both Table \ref{t3} and \ref{t4} are similar. When switching from AllCNN to ResNet, the vector-based framework observes performance drop at the highest measurement rate ($0.25$), but increases in lower rates ($0.1$ and $0.018$). For our framework when switching from AllCNN to ResNet, the test accuracies stay approximately similar or improve. 

Table \ref{t5} shows the empirical complexity of both frameworks with respect to different measurement configurations, excluding the base classifiers. Since all three datasets employed in this experiment have the same input size and the size of the feature tensor $\mathcal{T}$ in MCL was set similar to the original input size, the complexities of CS and FS components in all three datasets are similar. It is clear that our proposed MCL framework has much lower memory and computational complexity compared to the vector-based counterpart. \textit{In our proposed framework, even operating at the highest measurement rate $0.247$, the CS and FS components require only $2.5\mathrm{K}$ parameters and $5\mathrm{K}$ FLOPs, which are approximately $20$ times fewer than that of the vector-based framework operating at the lowest measurement rate $0.018$}. Interestingly, the optimal configuration at each measurement rate obtained in our framework also has lower or similar complexity than the other configuration. 

\begin{figure*}[]
	
	\centering
	\includegraphics[width=\textwidth]{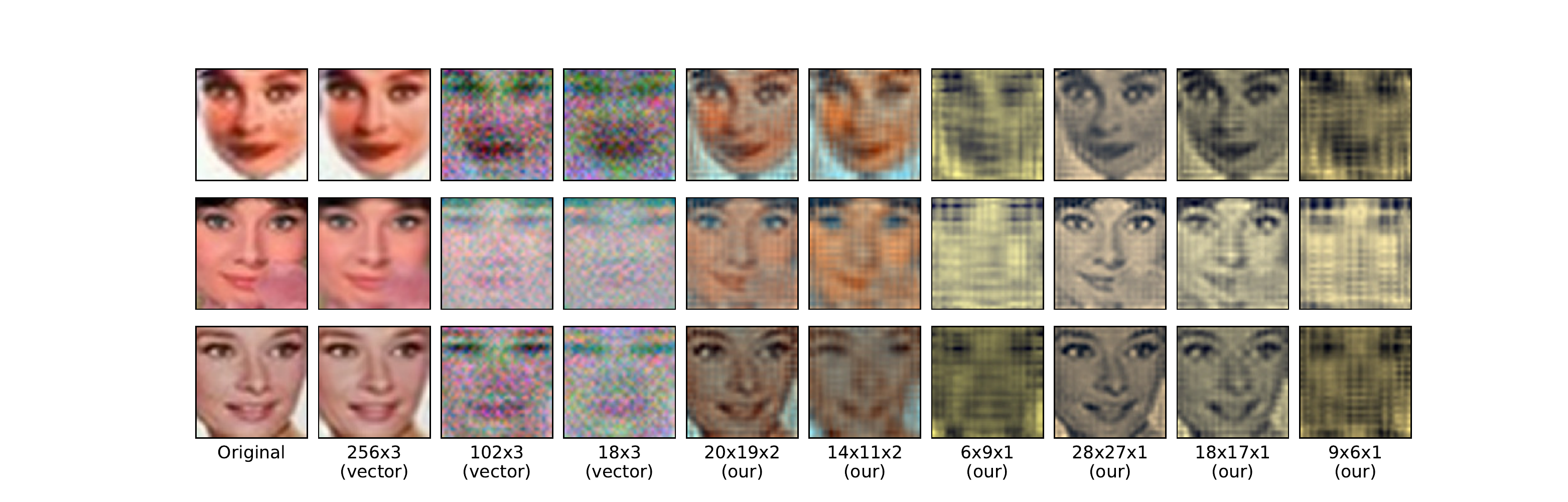}
	\caption{Illustration of the feature images (inputs to ResNet) synthesized by the proposed framework and the vector-based counterpart. The original images come from the test set of CelebA-32.}\label{figure2}
\end{figure*}

In Figure \ref{figure2}, we visualize the features obtained from the reprojection step and the FS component in the proposed framework, respectively. It is worth noting that the sensing matrices and the reprojection matrices (in case of the vector-based framework) or $\mathbf{\Theta}_n$ (in FS component of MCL framework) were initialized with PCA and HOSVD. In addition, the base network classifiers were also initialized with the ones trained on the original data. Thus, it is intuitive to expect the features obtained from both frameworks to be visually interpretable for human, despite no explicit reconstruction objective was incorporated during the training phase. Indeed, from Figure \ref{figure2}, we can see that with the highest number of measurements, the feature images obtained from both frameworks look very similar to the original images. Particularly, the ones synthesized by the vector-based framework look visually closer to the original images than those obtained from our MCL framework. Since the sensing and reprojection steps in the vector-based framework share the same weight matrices during the optimization procedure, the whole pipeline is more constrained to reconstruct the images at the reprojection step.

When the number of measurements drops to approximately $10\%$ of the original signal, the reverse scenario happens: the feature images (in configuration $14\times 11\times 2$, $28\times 27\times 1$) obtained from our framework retain more facial features compared to those from the vector-based framework ($102\times 3$), especially in the $28\times 27\times 1$ configuration. This is due to the fact that most of the facial information in particular, and natural images in general lie on the spatial dimensions, i.e., height and width. Besides, when the dimension of the third mode of the measurement $\mathcal{Z}$ is set to $1$ (as in configuration $28\times 27\times 1$, $18\times 17\times 1$), after the optimization procedure, our proposed framework effectively discards the color information which is less relevant to the facial recognition task, and retains more lightness details, thus, performs better than the configurations with the $3$-mode dimension set to $2$ (in configuration $20\times 19\times 2$, $14\times 11\times 2$). 

With the above observations from the empirical analysis, it is clear that structure-preserving Compressive Sensing and Feature Synthesis components in our proposed MCL framework can better capture essential information inherent in the multi-dimensional signal for the learning tasks, compared with the vector-based framework.  

\begin{table}[]
	\begin{center}
		\caption{Test Accuracy with AllCNN architecture as the base classifier}\label{t3}
		\resizebox{\linewidth}{!}{
			\begin{tabular}{|c|c|c|c|}\hline
				
Configuration 					& CIFAR-10 						& CIFAR-100 					& CelebA-32 \\ \hline \hline
Oracle							& $92.33$						& $72.25$						& $92.58$ \\ \hline \hline
$256 \times 3$ \cite{zisselman2018compressed}				& $81.36 \pm 00.00$				& $55.32 \pm 00.00$				& $89.25 \pm 00.00$ \\ \hline
$20\times 19\times 2$ (our)		& $\mathbf{89.35} \pm 00.26$	& $\mathbf{65.97} \pm 00.19$	& $92.36 \pm 00.07$ \\ \hline
$28\times 27\times 1$ (our)		& $88.56 \pm 00.14$				& $62.82 \pm 00.09$				& $\mathbf{92.74} \pm 00.31$ \\ \hline \hline
				
$102 \times 3$ \cite{zisselman2018compressed}				& $65.14 \pm 02.37$				& $44.03 \pm 03.60$				& $67.04 \pm 02.36$	\\ \hline
$14\times 11\times 2$ (our)		& $\mathbf{84.15} \pm 00.55$	& $\mathbf{59.77} \pm 00.12$	& $87.01 \pm 00.99$ \\ \hline
$18\times 17\times 1$ (our)		& $83.17 \pm 00.32$				& $54.96 \pm 00.17$				& $\mathbf{91.26} \pm 00.13$ \\ \hline \hline 
				
$18 \times 3$ \cite{zisselman2018compressed}				& $61.38 \pm 00.05$				& $\mathbf{37.78} \pm 00.08$	& $63.76 \pm 00.22$ \\ \hline
$9\times 6\times 1$ (our)		& $\mathbf{64.45} \pm 00.39$	& $34.74 \pm 00.29$				& $\mathbf{68.49} \pm 00.28$ \\ \hline 
$6\times 9\times 1$	(our)		& $64.28 \pm 00.35$				& $35.16 \pm 00.16$				& $65.92 \pm 00.77$ \\ \hline

			\end{tabular}
		}
	\end{center}
\end{table}

\begin{table}[]
	\begin{center}
		\caption{Test Accuracy with ResNet architecture as the base classifier}\label{t4}
		\resizebox{\linewidth}{!}{
			\begin{tabular}{|c|c|c|c|}\hline
				
				Configuration 					& CIFAR-10 						& CIFAR-100 					& CelebA-32 \\ \hline \hline 
				Oracle 							& $92.47$						& $72.38$						& $93.08$ \\ \hline \hline
				$256 \times 3$ \cite{zisselman2018compressed}				& $78.56 \pm 00.00$				& $53.03 \pm 00.00$				& $87.29 \pm 00.00$ \\ \hline
				$20\times 19\times 2$ (our)		& $\mathbf{89.22} \pm 00.27$	& $\mathbf{67.21} \pm 00.18$				& $92.00 \pm 00.48$ \\ \hline
				$28\times 27\times 1$ (our)		& $88.24 \pm 00.15$				& $63.37 \pm 00.44$				& $\mathbf{93.54} \pm 00.32$ \\ \hline \hline
				
				$102 \times 3$ \cite{zisselman2018compressed}				& $67.65 \pm 02.99$				& $47.90 \pm 01.22$				& $76.32 \pm 02.35$	\\ \hline
				$14\times 11\times 2$ (our)		& $\mathbf{84.74} \pm 00.16$	& $\mathbf{60.30} \pm 00.21$				& $88.50 \pm 00.26$ \\ \hline
				$18\times 17\times 1$ (our)		& $83.31 \pm 00.21$				& $55.51 \pm 00.08$				& $\mathbf{90.82} \pm 00.14$ \\ \hline \hline 
				
				$18 \times 3$ \cite{zisselman2018compressed}				& $61.96 \pm 00.17$				& $\mathbf{41.03} \pm 00.22$				& $67.29 \pm 00.44$ \\ \hline
				$9\times 6\times 1$ (our)		& $\mathbf{64.14} \pm 00.24$	& $33.67 \pm 00.17$				& $\mathbf{69.90} \pm 00.41$ \\ \hline 
				$6\times 9\times 1$	(our)		& $64.07 \pm 00.16$				& $32.40 \pm 01.62$				& $67.39 \pm 00.34$ \\ \hline

			\end{tabular}
		}
	\end{center}
\end{table}

\begin{table}[]
	\begin{center}
		\caption{Complexity of the proposed framework and the vector-based counterpart, excluding the base classifier component}\label{t5}
		\resizebox{0.7\linewidth}{!}{
			\begin{tabular}{|c|c|c|}\hline
				
				Configuration 					& \#Parameters 						& \#FLOPs 					 \\ \hline \hline 

				$256 \times 3$ \cite{zisselman2018compressed}			& $786\mathrm{K}$				& $1573\mathrm{K}$				\\ \hline
				$20\times 19\times 2$ (our)		& $\mathbf{2.5}\mathrm{K}$		& $\mathbf{5}\mathrm{K}$ \\ \hline
				$28\times 27\times 1$ (our)		& $3.5\mathrm{K}$	& $7\mathrm{K}$ \\ \hline \hline
				
				$102 \times 3$ \cite{zisselman2018compressed}				& $313\mathrm{K}$	& $627\mathrm{K}$ \\ \hline
				$14\times 11\times 2$ (our)		& $\mathbf{1.6}\mathrm{K}$	& $\mathbf{3.2}\mathrm{K}$ \\ \hline
				$18\times 17\times 1$ (our)		& $2.2\mathrm{K}$ 	& $4.5\mathrm{K}$ \\ \hline \hline
				
				$18 \times 3$ \cite{zisselman2018compressed}			& $55\mathrm{K}$			& $111\mathrm{K}$	\\ \hline
				$9\times 6\times 1$ (our)		& $\mathbf{1.0}\mathrm{K}$	& $\mathbf{1.9}\mathrm{K}$ \\ \hline
				$6\times 9\times 1$	(our)		& $\mathbf{1.0}\mathrm{K}$	& $\mathbf{1.9}\mathrm{K}$ \\ \hline

			\end{tabular}
		}
	\end{center}
\end{table}

\subsection{Ablation Study}

In this subsection, we provide the empirical analysis on the effect of different components in MCL framework. These factors include the effect of the popular nonlinear thresholding step discussed in Section III.B; the choice of having shared or separate weights in CS and FS component; the initialization step discussed in Section III.C; the scalability of the proposed framework when the original dimensionalities of the signal increase. Since the total number of experiment settings when combining all of the aforementioned factors is huge, and the results involved multiple factors are difficult to interpret, we analyze these factors in a progressive manner. 

\subsubsection{Linearity versus Nonlinearity and Shared versus Separate Weights}

Firstly, the choice of linearity or nonlinearity and the choice of shared or separate weights in CS and FS component are analyzed together since the two factors are closely related. In this setting, the CS and FS components are initialized by HOSVD decomposition as described in Section III.C. The neural network classifier $\mathfrak{N}$ has the AllCNN architecture with the weights initialized from the corresponding pre-trained network on the original data. Table \ref{t6} shows the test accuracies on CIFAR-10, CIFAR-100 and CelebA-32 at different measurements. It is clear that most of the highest test accuracies are obtained without the thresholding step and with separate weights in CS and FS component, i.e., most bold-face numbers appear in the lower quarter on the left side of Table \ref{t6}. Comparing between linearity and nonlinearity option, it is obvious that the nonlinearity effect of $\mathrm{ReLU}$ adversely affect the performances, especially when the number of measurements decreases. The reason might be that applying $\mathrm{ReLU}$ to the compressed measurements restricts the information to be represented in the positive subspace only, thus further losing the representation power in the compressed measurements when only a limited number of measurements allowed.  

In the linearity setting, while the performance differences between shared and separate weights in some configurations are small, here we should note that allowing non-shared weights can be beneficial in cases where we know that certain features should be synthesized in the FS component in order to make inferences. 

\subsubsection{Effect of The Initialization Step}

From the observation obtained from the above analysis on the effect of linearity and separate weights, we investigated the effect of the initialization step discussed in Section III.C. All setups were trained with a multilinear FS component having separate weights from CS component. From Table \ref{t7}, we can easily observe that by initializing the CS and FS components with HOSVD, the performances of the learning systems increase significantly. When CS and FS components are initialized with HOSVD, utilizing a pre-trained network further improves the inference performance of the systems, especially in the low measurement rate regime. Thus, the initialization strategy proposed in Section III.C is beneficial in a general setting for the learning tasks. 

\begin{table*}[]
	\begin{center}
		\caption{Test accuracy with respect to the choice of linearity or nonlinearity in conjunction with the choice of shared or separate weights in CS and FS component. The \textbf{bold} numbers denote the best test accuracy (among 4 combinations of \textit{LINEARITY} versus \textit{NONLINEARITY} and \textit{SHARED} versus \textit{SEPARATE}) in the same dataset with the same configuration}\label{t6}
		\resizebox{\textwidth}{!}{
			
\begin{tabular}{|c|c|c|c|c|c|c|c|}
	\hline
	\mr{2}{*}{} & \mr{2}{*}{Configuration} & \mc{3}{c|}{LINEARITY} & \mc{3}{c|}{NONLINEARITY}   \\\cl{3-8}
	&                 & CIFAR-10 & CIFAR-100 & CelebA-32 & CIFAR-10 & CIFAR-100 & CelebA-32 \\ \hline
	\mr{6}{*}{SHARED}            & $20\times 19\times 2$ & $89.25 \pm 00.39$ & $66.00\pm 00.26$ & $92.24\pm 00.21$ & $89.16\pm 00.14$ & $\mathbf{66.11}\pm 00.10$ & $91.69\pm 00.46$ \\\cl{2-8}
								 & $28\times 27\times 1$ & $88.44 \pm 00.04$ & $62.62\pm 00.30$ & $92.42\pm 00.64$ & $88.32\pm 00.20$ & $61.99\pm 01.24$ & $92.61\pm 00.61$ \\\cl{2-8}
								 & $14\times 11\times 2$ & $\mathbf{84.37}\pm 00.35$ & $59.75\pm 00.23$ & $\mathbf{88.00}\pm 00.38$ & $83.84\pm 00.21$ & $58.01\pm 00.80$ & $84.75\pm 01.37$ \\\cl{2-8}
								 & $18\times 17\times 1$ & $\mathbf{83.73}\pm 00.10$ & $54.88\pm 00.23$ & $91.22\pm 00.43$ & $83.57\pm 00.30$ & $54.47\pm 00.39$ & $90.85\pm 00.60$ \\\cl{2-8}
								 & $6\times 9\times 1$ & $64.27\pm 00.28$ & $33.17\pm 00.20$ & $64.47\pm 00.61$ & $62.18\pm 01.37$ & $33.04\pm 00.08$ & $49.46\pm 00.95$ \\\cl{2-8}
								 & $9\times 6\times 1$ & $64.43\pm 00.25$ & $32.82\pm 00.39$ & $68.25\pm 00.09$ & $62.34\pm 00.16$ & $33.08\pm 00.16$ & $59.29\pm 01.83$ \\ \hline \hline
	\mr{6}{*}{SEPARATE}          & $20\times 19\times 2$ & $\mathbf{89.35}\pm 00.26$ & $65.97\pm 00.19$ & $92.36\pm 00.07$ & $88.19\pm 01.00$ & $64.28\pm 02.29$ & $91.79\pm 00.21$ \\\cl{2-8}
	
								& $28\times 27\times 1$ & $\mathbf{88.56}\pm 00.14$ & $\mathbf{62.82}\pm 00.09$ & $\mathbf{92.74}\pm 00.31$ & $88.14\pm 00.34$ & $62.15\pm 00.04$ & $92.49\pm 00.43$ \\\cl{2-8}
								& $14\times 11\times 2$ & $84.15\pm 00.55$ & $\mathbf{59.77}\pm 00.12$ & $87.01\pm 00.99$ & $82.16\pm 02.61$ & $59.15\pm 00.16$ & $84.64\pm 00.57$ \\\cl{2-8}
								& $18\times 17\times 1$ & $83.17\pm 00.32$ & $\mathbf{54.96}\pm 00.17$ & $\mathbf{91.26}	\pm 00.14$ & $82.90\pm 00.19$ & $53.90\pm 00.41$ & $90.10\pm 00.69$ \\\cl{2-8}
								& $6\times 9\times 1$ & $\mathbf{64.28}\pm 00.35$ & $\mathbf{35.16}\pm 00.16$ & $\mathbf{65.92}\pm 00.77$ & $61.09\pm 00.20$ & $32.15\pm 00.49$ & $50.68\pm 01.00$ \\\cl{2-8}
								& $9\times 6\times 1$ & $\mathbf{64.45}\pm 00.39$ & $34.74\pm 00.29$ & $68.49\pm 00.28$ & $62.18\pm 00.25$ & $32.68\pm 00.81$ & $61.54\pm 01.40$ \\\cl{1-8}

\end{tabular}
		}
	\end{center}
\end{table*}

\begin{table*}[]
	\begin{center}
		\caption{Test accuracy with respect to the initialization of CS \& FS component and the base classifier (AllCNN). The \textbf{bold} numbers denote the best test accuracy (among 4 combinations of \textit{PRECOMPUTE CLASSIFIER} versus \textit{RANDOM CLASSIFIER} and \textit{PRECOMPUTE CS \& FS} versus \textit{RANDOM CS \& FS}) in the same dataset with the same configuration}\label{t7}
		\resizebox{\textwidth}{!}{
			
			\begin{tabular}{|c|c|c|c|c|c|c|c|}
				\hline
				\mr{2}{*}{} & \mr{2}{*}{Configuration} & \mc{3}{c|}{PRECOMPUTE CS \& FS} & \mc{3}{c|}{RANDOM CS \& FS}   \\\cl{3-8}
				&                 & CIFAR-10 & CIFAR-100 & CelebA-32 & CIFAR-10 & CIFAR-100 & CelebA-32 \\ \hline
				\mr{3}{*}{\makecell{PRECOMPUTE \\ CLASSIFIER}}            & $28\times 27\times 1$ & $88.56\pm 00.14$ & $\mathbf{62.82}\pm 00.09$ & $\mathbf{92.74}\pm 00.31$ & $71.47\pm 00.62$ & $38.59\pm 02.60$ & $68.65\pm 01.56$ \\\cl{2-8}
				& $18\times 17\times 1$  & $83.17\pm 00.32$ & $\mathbf{54.96}\pm 00.17$ & $\mathbf{91.26}\pm 00.14$ & $69.46\pm 01.16$ & $38.65\pm 00.28$ & $65.65\pm 00.70$ \\\cl{2-8}
				& $9\times 6\times 1$ & $\mathbf{64.45}\pm 00.39$ & $\mathbf{34.74}\pm 00.29$ & $\mathbf{68.49}\pm 00.28$ & $59.88\pm 00.17$ & $29.03\pm 01.69$ & $60.43\pm 02.29$  \\ \hline \hline
				\mr{3}{*}{\makecell{RANDOM \\ CLASSIFIER}} & $28\times 27\times 1$ & $\mathbf{88.71}\pm 00.05$ & $60.36\pm 00.62$ & $85.56\pm 00.84$ & $71.37\pm 01.24$ & $38.98\pm 01.40$ & $67.18\pm 00.32$ \\\cl{2-8}
				& $18\times 17\times 1$ & $\mathbf{83.82}\pm 00.28$ & $49.41\pm 04.07$ & $84.71\pm 01.39$ & $68.93\pm 01.16$ & $37.84\pm 03.06$ & $64.12\pm 02.06$ \\\cl{2-8}
				& $9\times 6\times 1$ & $63.99\pm 00.11$ & $33.66\pm 00.66$ & $65.62\pm 02.71$ & $59.85\pm 00.67$ & $30.90\pm 00.17$ & $56.61\pm 02.27$ \\ \hline
				
			\end{tabular}
		}
	\end{center}
\end{table*}

\subsubsection{Scalability}

Finally, the scalability of the proposed framework is validated in different resolutions of the CelebA dataset. All of the previous experiments were demonstrated with CelebA-32 dataset, which we assume that there are only $3072$ elements in the original signal. To investigate the scalability, we pose the following question: \textit{What if the original dimensions of the signal are higher than $32\times 32\times 3$, with the same numbers of measurements presented in Table \ref{t2}, can we still learn to recognize facial images with feasible costs?}. To answer this question, we trained our framework on CelebA-32, CelebA-48, CelebA-64 and CelebA-80 and recorded the test accuracies, the number of parameters and the number of FLOPs at different number of measurements, which are shown in Table \ref{t8}. It is clear that at each measurement configuration, when the original signal resolution increases, the measurement rate drops at a similar rate, however, without any adverse effect on the inference performance. Particularly, if we look into the last column of Table \ref{t8}, with a sampling rate of only $4\%$, the proposed framework achieves $93\%$ accuracy, which is only $2\%$ lower compared to that of the base classifier trained on the original data. Here we should note that most of the images in CelebA dataset have higher resolution than $80\times 80\times 3$ pixel, therefore, 4 different versions of CelebA ($32\times 32\times 3$, $48\times 48\times 3$, $64\times 64\times 3$, $80\times 80\times 3$, ) in our experiments indeed contain increasing levels of data fidelity. From the performance statistics, we can observe that the performance of our framework is characterized by the number of measurements, rather than the measurement rates or compression rates. 

\begin{figure*}[]
	
	\centering
	\includegraphics[width=\textwidth]{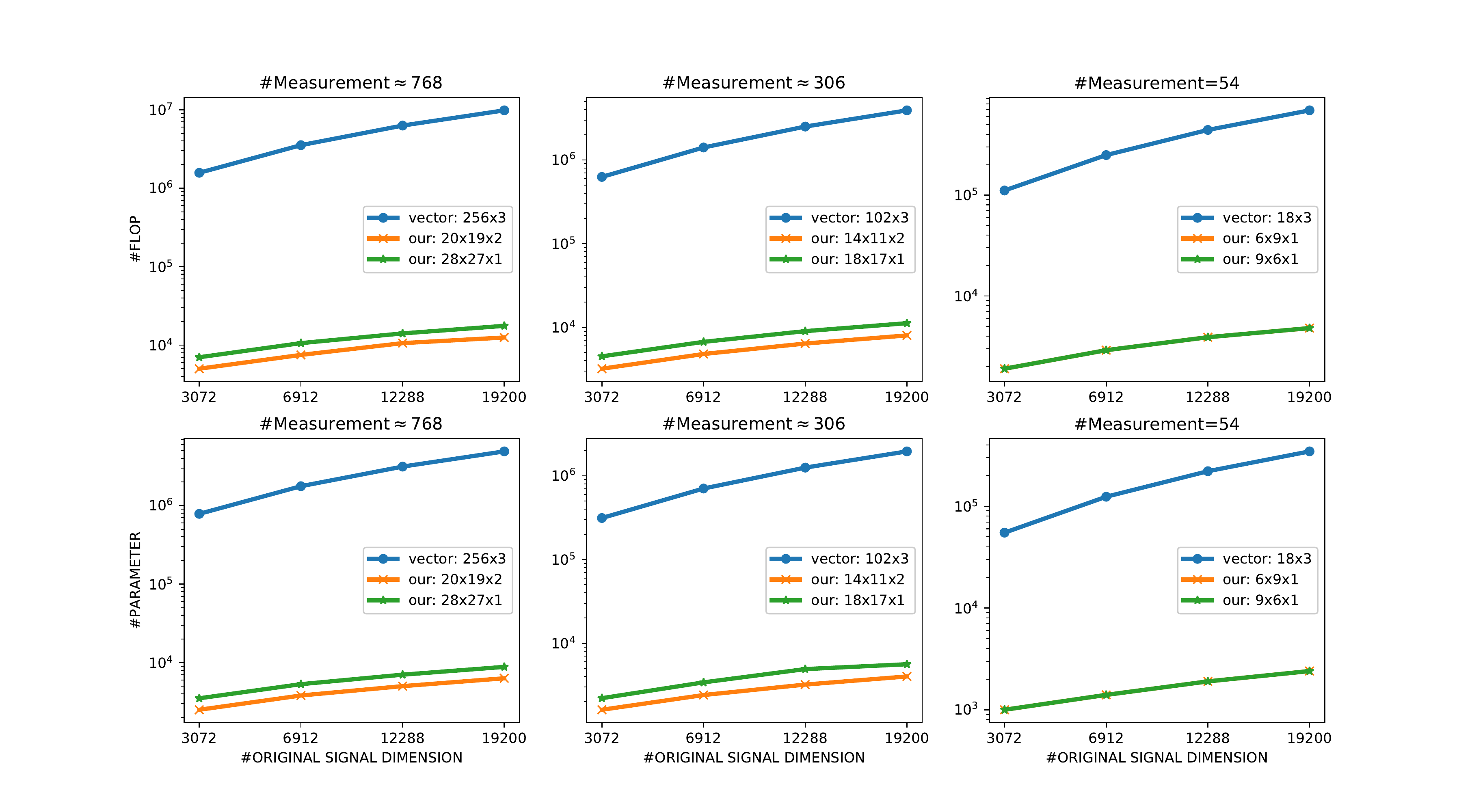}
	\caption{\#FLOP and \#PARAMETER versus the original dimensionalities of the signal, measured in the proposed framework and the vector-based framework, excluding the base classifier. The x-axis represents the original dimension of the input signal. The y-axis on the first row represents the number of FLOPs in log scale while the y-axis on the second row represents the number of parameters}\label{figure3}
\end{figure*}

Due to the memory limitation when training the vector-based framework at higher resolutions, we could not perform the same set of experiments for the vector-based framework. However, to compare the scalability in terms of computation and memory between the two frameworks, we measured the number of FLOPs and parameters in the vector-based framework, excluding the base classifier and visualize the results on Figure \ref{figure3}. It is worth noting that on the y-axis is the log scale and as the dimensions of the original signal increase, the complexity of the vector-based framework increases by an order of magnitude while our proposed MCL framework scales favorably in both memory and computation. 

\begin{table*}[]
	\begin{center}
		\caption{Test Performance \& Complexity of the proposed framework at different resolutions of the original CelebA dataset, with AllCNN as the base classifier}\label{t8}
		\resizebox{\textwidth}{!}{
			
			\begin{tabular}{|c|c|c|c|c|c|c|c|c|}
				\hline
\mr{2}{*}{Configuration} & \mc{4}{c|}{ACCURACY} & \mc{4}{c|}{MEASUREMENT RATE}   \\\cl{2-9}
& CelebA-32 & CelebA-48 & CelebA-64 & CelebA-80 & CelebA-32 & CelebA-48 & CelebA-64 & CelebA-80 \\ \hline
Oracle & $92.58$ & $93.37$ & $94.75$ & $95.04$ & $1.0$ & $1.0$ & $1.0$ & $1.0$ \\\cl{1-9}
$20\times 19\times 2$ & $92.36\pm 00.07$ & $91.24\pm 00.23$ & $92.62\pm 00.36$ & $93.01\pm 00.32$ & $0.247$ & $0.110$ & $0.062$ & $0.040$ \\ \hline
$28\times 27\times 1$ & $92.74\pm 00.31$ & $92.43\pm 00.19$ & $93.39\pm 00.52$ & $93.42\pm 00.37$ & $0.246$ & $0.109$ & $0.062$ & $0.039$ \\ \hline \hline
$14\times 11\times 2$ & $87.01\pm 00.99$ & $87.00\pm 00.87$ & $87.75\pm 00.27$ & $88.67\pm 00.26$ & $0.100$ & $0.045$ & $0.025$ & $0.016$ \\ \hline 
$18\times 17\times 1$ & $91.26\pm 00.14$ & $90.17\pm 00.30$ & $91.36\pm 00.43$ & $91.89\pm 00.46$ & $0.1$ & $0.044$ & $0.025$ & $0.016$ \\ \hline \hline
$6\times 9\times 1$ & $65.92\pm 00.77$ & $66.56\pm 00.05$ & $66.69\pm 00.46$ & $66.03\pm 00.37$ & $0.018$ & $0.008$ & $0.004$ & $0.003$ \\ \hline
$9\times 6\times 1$ & $68.49\pm 00.28$ & $68.69\pm 00.96$ & $67.75\pm 01.62$ & $67.31\pm 00.49$ & $0.018$ & $0.008$ & $0.004$ & $0.003$ \\ \hline \hline
\mr{2}{*}{Configuration} & \mc{4}{c|}{\#FLOP} & \mc{4}{c|}{\#PARAMETER}   \\\cl{2-9}
& CelebA-32 & CelebA-48 & CelebA-64 & CelebA-80 & CelebA-32 & CelebA-48 & CelebA-64 & CelebA-80 \\ \hline
$20\times 19\times 2$ & $50\mathrm{K}$ & $7.5\mathrm{K}$ & $10.0\mathrm{K}$ & $12.5\mathrm{K}$ & $2.5\mathrm{K}$ & $3.8\mathrm{K}$ & $5.0\mathrm{K}$ & $6.3\mathrm{K}$ \\\cl{1-9}
$28\times 27\times 1$ & $7.0\mathrm{K}$ & $10.6\mathrm{K}$ & $14.1\mathrm{K}$ & $17.6\mathrm{K}$ & $3.5\mathrm{K}$ & $5.3\mathrm{K}$ & $7.0\mathrm{K}$ & $8.8\mathrm{K}$ \\ \hline \hline
$14\times 11\times 2$ & $3.2\mathrm{K}$ & $4.8\mathrm{K}$ & $6.4\mathrm{K}$ & $8.0\mathrm{K}$ & $1.6\mathrm{K}$ & $2.4\mathrm{K}$ & $3.2\mathrm{K}$ & $4.0\mathrm{K}$ \\ \hline 
$18\times 17\times 1$ & $4.5\mathrm{K}$ & $6.7\mathrm{K}$ & $9.0\mathrm{K}$ & $11.2\mathrm{K}$ & $2.2\mathrm{K}$ & $3.4\mathrm{K}$ & $4.9\mathrm{K}$ & $5.6\mathrm{K}$ \\ \hline \hline
$6\times 9\times 1$ & $1.9\mathrm{K}$ & $2.9\mathrm{K}$ & $3.9\mathrm{K}$ & $4.8\mathrm{K}$ & $1.0\mathrm{K}$ & $1.4\mathrm{K}$ & $1.9\mathrm{K}$ & $2.4\mathrm{K}$ \\ \hline
$9\times 6\times 1$ & $1.9\mathrm{K}$ & $2.9\mathrm{K}$ & $3.9\mathrm{K}$ & $4.8\mathrm{K}$ & $1.0\mathrm{K}$ & $1.4\mathrm{K}$ & $1.9\mathrm{K}$ & $2.4\mathrm{K}$ \\ \hline
			\end{tabular}
		}
	\end{center}
\end{table*}

\section{Conclusions}
In this paper, we proposed Multilinear Compressive Learning, an efficient framework to tackle the Compressive Learning task that operates on multi-dimensional signals. The proposed framework takes into account the tensorial nature of the multi-dimensional signals and performs the compressive sensing as well as the feature extraction step along different modes of the original data, thus being able to retain and synthesize essential information on a multilinear subspace for the learning task. We show theoretically and empirically that the proposed framework outperforms its vector-based counterpart in both inference performance and computational efficiency. Extensive ablation study has been conducted to investigate the effect of different components in the proposed framework, giving insights into the importance of different design choices. 

\section{Acknowledgement}
This project has received funding from the European Union's Horizon 2020 research and innovation programme under grant agreement No 871449 (OpenDR). This publication reflects the authors’ views only. The European Commission is not responsible for any use that may be made of the information it contains.

\bibliography{manuscript}

\begin{thebibliography}{10}

\bibitem{candes2008introduction}
E.~J. Cand{\`e}s and M.~B. Wakin, ``An introduction to compressive sampling [a
  sensing/sampling paradigm that goes against the common knowledge in data
  acquisition],'' {\em IEEE signal processing magazine}, vol.~25, no.~2,
  pp.~21--30, 2008.

\bibitem{candes2006stable}
E.~J. Candes, J.~K. Romberg, and T.~Tao, ``Stable signal recovery from
  incomplete and inaccurate measurements,'' {\em Communications on Pure and
  Applied Mathematics: A Journal Issued by the Courant Institute of
  Mathematical Sciences}, vol.~59, no.~8, pp.~1207--1223, 2006.

\bibitem{donoho2006compressed}
D.~L. Donoho {\em et~al.}, ``Compressed sensing,'' {\em IEEE Transactions on
  information theory}, vol.~52, no.~4, pp.~1289--1306, 2006.

\bibitem{mohassel2017secureml}
P.~Mohassel and Y.~Zhang, ``Secureml: A system for scalable privacy-preserving
  machine learning,'' in {\em 2017 IEEE Symposium on Security and Privacy
  (SP)}, pp.~19--38, IEEE, 2017.

\bibitem{hesamifard2017cryptodl}
E.~Hesamifard, H.~Takabi, and M.~Ghasemi, ``Cryptodl: Deep neural networks over
  encrypted data,'' {\em arXiv preprint arXiv:1711.05189}, 2017.

\bibitem{calderbank2012finding}
R.~Calderbank and S.~Jafarpour, ``Finding needles in compressed haystacks,'' in
  {\em 2012 IEEE International Conference on Acoustics, Speech and Signal
  Processing (ICASSP)}, pp.~3441--3444, IEEE, 2012.

\bibitem{davenport2007smashed}
M.~A. Davenport, M.~F. Duarte, M.~B. Wakin, J.~N. Laska, D.~Takhar, K.~F.
  Kelly, and R.~G. Baraniuk, ``The smashed filter for compressive
  classification and target recognition,'' in {\em Computational Imaging V},
  vol.~6498, p.~64980H, International Society for Optics and Photonics, 2007.

\bibitem{davenport2010signal}
M.~A. Davenport, P.~Boufounos, M.~B. Wakin, R.~G. Baraniuk, {\em et~al.},
  ``Signal processing with compressive measurements.,'' {\em J. Sel. Topics
  Signal Processing}, vol.~4, no.~2, pp.~445--460, 2010.

\bibitem{reboredo2013compressive}
H.~Reboredo, F.~Renna, R.~Calderbank, and M.~R. Rodrigues, ``Compressive
  classification,'' in {\em 2013 IEEE International Symposium on Information
  Theory}, pp.~674--678, IEEE, 2013.

\bibitem{baheti2008adaptive}
P.~K. Baheti and M.~A. Neifeld, ``Adaptive feature-specific imaging: a face
  recognition example,'' {\em Applied optics}, vol.~47, no.~10, pp.~B21--B31,
  2008.

\bibitem{reboredo2013projections}
H.~Reboredo, F.~Renna, R.~Calderbank, and M.~R. Rodrigues, ``Projections
  designs for compressive classification,'' in {\em 2013 IEEE Global Conference
  on Signal and Information Processing}, pp.~1029--1032, IEEE, 2013.

\bibitem{lohit2015reconstruction}
S.~Lohit, K.~Kulkarni, P.~Turaga, J.~Wang, and A.~C. Sankaranarayanan,
  ``Reconstruction-free inference on compressive measurements,'' in {\em
  Proceedings of the IEEE Conference on Computer Vision and Pattern Recognition
  Workshops}, pp.~16--24, 2015.

\bibitem{adler2016compressed}
A.~Adler, M.~Elad, and M.~Zibulevsky, ``Compressed learning: A deep neural
  network approach,'' {\em arXiv preprint arXiv:1610.09615}, 2016.

\bibitem{lohit2016direct}
S.~Lohit, K.~Kulkarni, and P.~Turaga, ``Direct inference on compressive
  measurements using convolutional neural networks,'' in {\em 2016 IEEE
  International Conference on Image Processing (ICIP)}, pp.~1913--1917, IEEE,
  2016.

\bibitem{nion2010tensor}
D.~Nion and N.~D. Sidiropoulos, ``Tensor algebra and multidimensional harmonic
  retrieval in signal processing for mimo radar,'' {\em IEEE Transactions on
  Signal Processing}, vol.~58, no.~11, pp.~5693--5705, 2010.

\bibitem{miwakeichi2004decomposing}
F.~Miwakeichi, E.~Mart{\i}nez-Montes, P.~A. Vald{\'e}s-Sosa, N.~Nishiyama,
  H.~Mizuhara, and Y.~Yamaguchi, ``Decomposing eeg data into
  space--time--frequency components using parallel factor analysis,'' {\em
  NeuroImage}, vol.~22, no.~3, pp.~1035--1045, 2004.

\bibitem{dunlavy2011multilinear}
D.~M. Dunlavy, T.~G. Kolda, and W.~P. Kegelmeyer, ``Multilinear algebra for
  analyzing data with multiple linkages,'' in {\em Graph algorithms in the
  language of linear algebra}, pp.~85--114, SIAM, 2011.

\bibitem{tran2017tensor}
D.~T. Tran, M.~Magris, J.~Kanniainen, M.~Gabbouj, and A.~Iosifidis, ``Tensor
  representation in high-frequency financial data for price change
  prediction,'' in {\em 2017 IEEE Symposium Series on Computational
  Intelligence (SSCI)}, pp.~1--7, IEEE, 2017.

\bibitem{tran2018improving}
D.~T. Tran, A.~Iosifidis, and M.~Gabbouj, ``Improving efficiency in
  convolutional neural networks with multilinear filters,'' {\em Neural
  Networks}, vol.~105, pp.~328--339, 2018.

\bibitem{cichocki2015tensor}
A.~Cichocki, D.~Mandic, L.~De~Lathauwer, G.~Zhou, Q.~Zhao, C.~Caiafa, and H.~A.
  Phan, ``Tensor decompositions for signal processing applications: From
  two-way to multiway component analysis,'' {\em IEEE Signal Processing
  Magazine}, vol.~32, no.~2, pp.~145--163, 2015.

\bibitem{malgouyres2018multilinear}
F.~Malgouyres and J.~Landsberg, ``Multilinear compressive sensing and an
  application to convolutional linear networks,'' 2018.

\bibitem{tran2018temporal}
D.~T. Tran, A.~Iosifidis, J.~Kanniainen, and M.~Gabbouj, ``Temporal
  attention-augmented bilinear network for financial time-series data
  analysis,'' {\em IEEE transactions on neural networks and learning systems},
  2018.

\bibitem{youd2002revised}
T.~L. Youd, C.~M. Hansen, and S.~F. Bartlett, ``Revised multilinear regression
  equations for prediction of lateral spread displacement,'' {\em Journal of
  Geotechnical and Geoenvironmental Engineering}, vol.~128, no.~12,
  pp.~1007--1017, 2002.

\bibitem{zhao2013higher}
Q.~Zhao, C.~F. Caiafa, D.~P. Mandic, Z.~C. Chao, Y.~Nagasaka, N.~Fujii,
  L.~Zhang, and A.~Cichocki, ``Higher order partial least squares (hopls): a
  generalized multilinear regression method,'' {\em IEEE transactions on
  pattern analysis and machine intelligence}, vol.~35, no.~7, pp.~1660--1673,
  2013.

\bibitem{li2014multilinear}
Q.~Li and D.~Schonfeld, ``Multilinear discriminant analysis for higher-order
  tensor data classification,'' {\em IEEE transactions on pattern analysis and
  machine intelligence}, vol.~36, no.~12, pp.~2524--2537, 2014.

\bibitem{tran2017multilinear}
D.~T. Tran, M.~Gabbouj, and A.~Iosifidis, ``Multilinear class-specific
  discriminant analysis,'' {\em Pattern Recognition Letters}, vol.~100,
  pp.~131--136, 2017.

\bibitem{denton2014exploiting}
E.~L. Denton, W.~Zaremba, J.~Bruna, Y.~LeCun, and R.~Fergus, ``Exploiting
  linear structure within convolutional networks for efficient evaluation,'' in
  {\em Advances in neural information processing systems}, pp.~1269--1277,
  2014.

\bibitem{jaderberg2014speeding}
M.~Jaderberg, A.~Vedaldi, and A.~Zisserman, ``Speeding up convolutional neural
  networks with low rank expansions,'' {\em arXiv preprint arXiv:1405.3866},
  2014.

\bibitem{lebedev2014speeding}
V.~Lebedev, Y.~Ganin, M.~Rakhuba, I.~Oseledets, and V.~Lempitsky, ``Speeding-up
  convolutional neural networks using fine-tuned cp-decomposition,'' {\em arXiv
  preprint arXiv:1412.6553}, 2014.

\bibitem{yang2017tensor}
Y.~Yang, D.~Krompass, and V.~Tresp, ``Tensor-train recurrent neural networks
  for video classification,'' in {\em Proceedings of the 34th International
  Conference on Machine Learning-Volume 70}, pp.~3891--3900, JMLR. org, 2017.

\bibitem{caiafa2013multidimensional}
C.~F. Caiafa and A.~Cichocki, ``Multidimensional compressed sensing and their
  applications,'' {\em Wiley Interdisciplinary Reviews: Data Mining and
  Knowledge Discovery}, vol.~3, no.~6, pp.~355--380, 2013.

\bibitem{yu2014multidimensional}
Y.~Yu, J.~Jin, F.~Liu, and S.~Crozier, ``Multidimensional compressed sensing
  mri using tensor decomposition-based sparsifying transform,'' {\em PloS one},
  vol.~9, no.~6, p.~e98441, 2014.

\bibitem{robucci2008compressive}
R.~Robucci, L.~K. Chiu, J.~Gray, J.~Romberg, P.~Hasler, and D.~Anderson,
  ``Compressive sensing on a cmos separable transform image sensor,'' in {\em
  2008 IEEE International Conference on Acoustics, Speech and Signal
  Processing}, pp.~5125--5128, IEEE, 2008.

\bibitem{aharon2006k}
M.~Aharon, M.~Elad, A.~Bruckstein, {\em et~al.}, ``K-svd: An algorithm for
  designing overcomplete dictionaries for sparse representation,'' {\em IEEE
  Transactions on signal processing}, vol.~54, no.~11, p.~4311, 2006.

\bibitem{donoho2003optimally}
D.~L. Donoho and M.~Elad, ``Optimally sparse representation in general
  (nonorthogonal) dictionaries via l1 minimization,'' {\em Proceedings of the
  National Academy of Sciences}, vol.~100, no.~5, pp.~2197--2202, 2003.

\bibitem{tropp2010computational}
J.~A. Tropp and S.~J. Wright, ``Computational methods for sparse solution of
  linear inverse problems,'' {\em Proceedings of the IEEE}, vol.~98, no.~6,
  pp.~948--958, 2010.

\bibitem{chen2001atomic}
S.~S. Chen, D.~L. Donoho, and M.~A. Saunders, ``Atomic decomposition by basis
  pursuit,'' {\em SIAM review}, vol.~43, no.~1, pp.~129--159, 2001.

\bibitem{tropp2004greed}
J.~A. Tropp, ``Greed is good: Algorithmic results for sparse approximation,''
  {\em IEEE Transactions on Information theory}, vol.~50, no.~10,
  pp.~2231--2242, 2004.

\bibitem{tropp2007signal}
J.~A. Tropp and A.~C. Gilbert, ``Signal recovery from random measurements via
  orthogonal matching pursuit,'' {\em IEEE Transactions on information theory},
  vol.~53, no.~12, pp.~4655--4666, 2007.

\bibitem{de2000multilinear}
L.~De~Lathauwer, B.~De~Moor, and J.~Vandewalle, ``A multilinear singular value
  decomposition,'' {\em SIAM journal on Matrix Analysis and Applications},
  vol.~21, no.~4, pp.~1253--1278, 2000.

\bibitem{duarte2012kronecker}
M.~F. Duarte and R.~G. Baraniuk, ``Kronecker compressive sensing,'' {\em IEEE
  Transactions on Image Processing}, vol.~21, no.~2, pp.~494--504, 2012.

\bibitem{caiafa2013computing}
C.~F. Caiafa and A.~Cichocki, ``Computing sparse representations of
  multidimensional signals using kronecker bases,'' {\em Neural computation},
  vol.~25, no.~1, pp.~186--220, 2013.

\bibitem{baraniuk2009random}
R.~G. Baraniuk and M.~B. Wakin, ``Random projections of smooth manifolds,''
  {\em Foundations of computational mathematics}, vol.~9, no.~1, pp.~51--77,
  2009.

\bibitem{kulkarni2012recurrence}
K.~Kulkarni and P.~Turaga, ``Recurrence textures for human activity recognition
  from compressive cameras,'' in {\em 2012 19th IEEE International Conference
  on Image Processing}, pp.~1417--1420, IEEE, 2012.

\bibitem{kulkarni2016reconstruction}
K.~Kulkarni and P.~Turaga, ``Reconstruction-free action inference from
  compressive imagers,'' {\em IEEE transactions on pattern analysis and machine
  intelligence}, vol.~38, no.~4, pp.~772--784, 2016.

\bibitem{hollis2018compressed}
B.~Hollis, S.~Patterson, and J.~Trinkle, ``Compressed learning for tactile
  object recognition,'' {\em IEEE Robotics and Automation Letters}, vol.~3,
  no.~3, pp.~1616--1623, 2018.

\bibitem{deugerli2018compressively}
A.~De{\u{g}}erli, S.~Aslan, M.~Yamac, B.~Sankur, and M.~Gabbouj,
  ``Compressively sensed image recognition,'' in {\em 2018 7th European
  Workshop on Visual Information Processing (EUVIP)}, pp.~1--6, IEEE, 2018.

\bibitem{xu2019compressed}
Y.~Xu and K.~F. Kelly, ``Compressed domain image classification using a
  multi-rate neural network,'' {\em arXiv preprint arXiv:1901.09983}, 2019.

\bibitem{glorot2010understanding}
X.~Glorot and Y.~Bengio, ``Understanding the difficulty of training deep
  feedforward neural networks,'' in {\em Proceedings of the thirteenth
  international conference on artificial intelligence and statistics},
  pp.~249--256, 2010.

\bibitem{he2016identity}
K.~He, X.~Zhang, S.~Ren, and J.~Sun, ``Identity mappings in deep residual
  networks,'' in {\em European conference on computer vision}, pp.~630--645,
  Springer, 2016.

\bibitem{wu2013supervised}
F.~Wu, X.~Tan, Y.~Yang, D.~Tao, S.~Tang, and Y.~Zhuang, ``Supervised
  nonnegative tensor factorization with maximum-margin constraint,'' in {\em
  Twenty-Seventh AAAI Conference on Artificial Intelligence}, 2013.

\bibitem{kim2007nonnegative}
Y.-D. Kim and S.~Choi, ``Nonnegative tucker decomposition,'' in {\em 2007 IEEE
  Conference on Computer Vision and Pattern Recognition}, pp.~1--8, IEEE, 2007.

\bibitem{grasedyck2013literature}
L.~Grasedyck, D.~Kressner, and C.~Tobler, ``A literature survey of low-rank
  tensor approximation techniques,'' {\em GAMM-Mitteilungen}, vol.~36, no.~1,
  pp.~53--78, 2013.

\bibitem{zisselman2018compressed}
E.~Zisselman, A.~Adler, and M.~Elad, ``Compressed learning for image
  classification: A deep neural network approach,'' {\em Processing, Analyzing
  and Learning of Images, Shapes, and Forms}, vol.~19, p.~1, 2018.

\bibitem{krizhevsky2009learning}
A.~Krizhevsky and G.~Hinton, ``Learning multiple layers of features from tiny
  images,'' tech. rep., Citeseer, 2009.

\bibitem{liu2015faceattributes}
Z.~Liu, P.~Luo, X.~Wang, and X.~Tang, ``Deep learning face attributes in the
  wild,'' in {\em Proceedings of International Conference on Computer Vision
  (ICCV)}, 2015.

\bibitem{springenberg2014striving}
J.~T. Springenberg, A.~Dosovitskiy, T.~Brox, and M.~Riedmiller, ``Striving for
  simplicity: The all convolutional net,'' {\em arXiv preprint
  arXiv:1412.6806}, 2014.

\bibitem{he2015delving}
K.~He, X.~Zhang, S.~Ren, and J.~Sun, ``Delving deep into rectifiers: Surpassing
  human-level performance on imagenet classification,'' in {\em Proceedings of
  the IEEE international conference on computer vision}, pp.~1026--1034, 2015.

\bibitem{kingma2014adam}
D.~P. Kingma and J.~Ba, ``Adam: A method for stochastic optimization,'' {\em
  arXiv preprint arXiv:1412.6980}, 2014.

\end{thebibliography}
\bibliographystyle{ieeetr}

\end{document}